\documentclass{article}
\usepackage[affil-it]{authblk}
\usepackage[bottom]{footmisc}

\usepackage{geometry}
\makeatletter
\def\@maketitle{%
  \newpage
  \null
  \vskip 2em%
  \begin{center}%
  \let \footnote \thanks
    {\Large\bfseries \@title \par}%
    \vskip 1.5em%
    {\normalsize
      \lineskip .5em%
      \begin{tabular}[t]{c}%
        \@author
      \end{tabular}\par}%
    \vskip 1.5em%
    {\normalsize \@date}%
  \end{center}%
  \par
  \vskip 1.5em}
\makeatother

\usepackage{doc}
\usepackage{subcaption}

\usepackage{url}

\usepackage{xcolor}

\usepackage{graphicx}
\usepackage{epstopdf}
\usepackage{amsmath}
\usepackage{amssymb}
\usepackage{bm}
\title{Comment on \textit{Adv-BNN: Improved Adversarial Defense through Robust Bayesian Neural Network}}

\author[1, 2]{Roland S. Zimmermann}

\affil[1]{Georg-August University of G\"ottingen, Germany}
\affil[2]{University of T\"ubingen, T\"ubingen, Germany}

\date{Dated: \today}

\begin{document}

\maketitle

\textbf{A recent paper \cite{liu2018adv} by Liu et al. combines the topics of adversarial training and Bayesian Neural Networks (BNN) and suggests that adversarially trained BNNs are more robust against adversarial attacks than their non-Bayesian counterparts. Here, I analyze the proposed defense and suggest that one needs to adjust the adversarial attack to incorporate the stochastic nature of a Bayesian network to perform an accurate evaluation of its robustness. Using this new type of attack I show that there appears to be no strong evidence for higher robustness of the adversarially trained BNNs.}\\

The evaluation of a neural network has proven to be a complex and difficult task, as one needs to separate two causes for the same observation - the robustness of the defended network and the shortcomings of the attack. If a network appears to be robust, this can either mean that it is in fact robust against adversarial attacks or that the attack is incomplete or relies on inapplicable assumptions on the attacked network.

Recently, a new paper \cite{liu2018adv} applied adversarial training \cite{madry2017towards} on Bayesian Neural Networks (BNNs). They call this combination Adv-BNN. The authors suggest that these models become more robust against $l_\infty$ adversarial attacks.  Here, I show that the reported robustness improvement is not based on an actually more robust network but on an insufficient attack algorithm. Furthermore, I show that one can in fact obtain a more robust model by using a sufficiently strong attack.

Liu et al. \cite{liu2018adv} introduce a type of BNN which learns not a point estimate of the weights but the parameters of a Gaussian distribution from which the weights are sampled during inference. Thus, the networks predictions are not deterministic but stochastic - this stochasticity can be expressed using a random vector $\bm{\epsilon}$ as an additional input to the network \cite{liu2018adv}. Therefore, to obtain a prediction Liu et al. average over multiple predictions of the network, i.e. multiple values of $\bm{\epsilon}$. By including this into the definition of an adversarial example $\hat{\mathbf{x}}$, they obtain
\begin{align}
    \hat{\mathbf{x}} = \underset{\mathbf{x} \in \mathit{S}}{\arg \max} \,\underset{\bm{\epsilon}}{\mathbb{E}}\left[ L(f(\mathbf{x}; \mathbf{w}, \bm{\epsilon}), \hat{y})\right].
\end{align}
Next, they notice that one cannot use the usual PGD algorithm \cite{madry2017towards} to find adversarial examples but need to take the stochasticity of the BNN into account. For this, they suggest to use a SGD formulation and come up with the update method
\begin{align}
    \label{eq:liu_bayesian_pgd}
    \hat{\mathbf{x}}_{t+1} \leftarrow \underset{\bm{x}+\mathit{S}}{\Pi} \left[ \hat{\mathbf{x}}_{t} + \eta \nabla_\mathbf{x}L\left(f(\bm{x}; \bm{w}, \bm{\epsilon}), \hat{y}\right)\Bigr\vert_{\substack{\mathbf{x} = \hat{\bm{x}}_{t} \\ \bm{\epsilon} = \bm{\epsilon}_{t}}} \right],
\end{align}
for which they sample a new random vector $\bm{\epsilon}$ in each update step. Here, $\Pi_{\bm{x}+\mathit{S}}$ represents the projection onto the set of allowed perturbations $\mathit{S}$, e.g. an $l_p$-ball around $\mathbf{x}$. In the following, this attack method will be called naive PGD. Just like the predictions of the networks its gradients are stochastic, too. Thus, the gradient direction from a single inference can point into an unrepresentative direction which makes the gradient ascent harder to perform.

To solve this issue I propose to use a modified version of their algorithm which comes closer to the original definition of the PGD attack. By estimating the real gradient of the network as the average of the gradients over multiple random vectors $\bm{\epsilon}$, one can obtain a more stable and therefore efficient attack
\begin{align}
    \label{eq:new_bayesian_pgd}
    \hat{\mathbf{x}}_{t+1} \leftarrow \underset{\bm{x}+\mathit{S}}{\Pi} \left[ \hat{\mathbf{x}}_{t} + \eta \underset{\mathbf{\epsilon}}{\mathbb{E}}\,\left( \nabla_\mathbf{x}L\left(f(\bm{x}; \bm{w}, \bm{\epsilon}), \hat{y}\right)\Bigr\vert_{\mathbf{x} = \hat{\bm{x}}_{t}} \right)\right],
\end{align}
which will be called averaged PGD (A-PGD) in the following. The reasoning behind this attack is similar to the motivation of the EOT attack \cite{athalye2018synthesizing}.

Based on the source code, instructions in their paper and the pre-trained models released by the authors I firstly reproduced the results reported for the Adv-BNN\footnote{For the BNNs I used an ensemble size of 40 during inference as suggested by Figure~4 in \cite{liu2018adv}.} on the STL-10 dataset \cite{coates2011analysis}. I did not evaluate the experiments on the CIFAR10 or ImageNet-143 datasets because of issues\footnote{https://github.com/xuanqing94/BayesianDefense/issues/5} in the published implementation. Therefore, my analysis is limited to the STL-10 dataset. I tested the defended networks on adversarial examples which are bounded by an $l_\infty$ ball of radius $0.035$ respectively $0.07$ and let the attacks run for $150$ steps \footnote{Increasing the number of attack steps further did not increase the attack success rate anymore.}; the results are shown in Table~\ref{table:accuracy}. Next, I used the A-PGD attack to generate the adversarial examples to test the robustness, which resulted in a decreased accuracy compared to the naive PGD. In the next step, I tweaked the hyperparameters of the adversarial training on the non-Bayesian network and obtained a model which was more robust than their reported baseline. Now, the Adv-BNN attacked by A-PGD was not significantly more robust, but even weaker for large adversarials than the non-Bayesian network. Finally, I used the A-PGD attack in the Adv-BNN method and obtained a network which was more robust than the other two.\\

\begin{table}[h]
\centering
\begin{tabular}{lccccc}
 \hline
 \multicolumn{2}{c}{} & \multicolumn{2}{c}{$\gamma=0.035$} & \multicolumn{2}{c}{$\gamma=0.07$} \\
 Defense & Plain & Naive PGD & A-PGD & Naive PGD & A-PGD \\
 \hline
 \hline
 Adv. Training     & 53.9\,\%           & 30.6\,\% & -                   & 10.8\,\% & - \\
 Adv-BNN           & \textbf{59.9}\,\%  & 37.6\,\% & 30.3\,\%            & 21.1\,\% & 8.6\,\% \\
 Adv-BNN w/ A-PGD  & 53.2\%             & 38.2\,\% & \textbf{31.2}\,\%   & 31.2\,\% & \textbf{13.4}\,\% \\
 \hline
\end{tabular}
\caption{\label{table:accuracy}Adversarial robustness of three different defense methods on the STL-10 dataset: an adversarially trained standard CNN (Adv. Training), the adversarially trained Bayesian variant of it proposed in \cite{liu2018adv} (Adv-BNN) and the Bayesian variant adversarially trained on A-PGD examples (Adv-BNN w/ A-PGD). The defended networks are tested on adversarial examples which are bounded by an $l_\infty$ ball of radius $\epsilon$ for $\gamma=0.035$ and $\gamma=0.07$.}
\end{table}

In conclusion, I have demonstrated that the Adv-BNN approach shows a performance increment tested on adversarial examples generated by the naive PGD method compared to the adversarial training of a standard CNN. But it's advantage vanishes if the modified A-PGD method is used to generate adversarial examples to test its robustness. If examples from the A-PGD attack are used to train the Adv-BNN, the performance increases again and yields a slightly higher accuracy than the adversarially trained standard CNN and the Adv-BNN based on the naive PGD attack, even though one can argue that this small advantage is not significant but merely based on the choice of hyperparamters. The findings show the importance of averaging the gradients in gradient-based attacks if Bayesian networks are targeted, as otherwise their robustness will be overestimated.

\bibliography{references}
\bibliographystyle{unsrt}

\end{document}